\documentclass[conference,twocolumn]{IEEEtran}
\IEEEoverridecommandlockouts

\usepackage{cite}
\usepackage{amsmath,amssymb,amsfonts}
\usepackage{algorithmic}
\usepackage{graphicx}
\usepackage{textcomp}
\usepackage{xcolor}
\usepackage{booktabs}
\usepackage{multirow}
\usepackage{array}
\usepackage{tabularx}
\usepackage{float}
\usepackage[pdfa]{hyperref}
\usepackage{balance}
\usepackage{stfloats}

\graphicspath{{img/}{figures/}}

\def\BibTeX{{\rm B\kern-.05em{\sc i\kern-.025em b}\kern-.08em
    T\kern-.1667em\lower.7ex\hbox{E}\kern-.125emX}}

\newcommand{\modelname}{\texorpdfstring{$\mu$Bi-ConvLSTM}{uBi-ConvLSTM}}
\newcommand{\modelnameplain}{\texorpdfstring{$\mu$Bi-ConvLSTM}{uBi-ConvLSTM}}

\begin{document}

\title{\texorpdfstring{$\mu$Bi-ConvLSTM}{uBi-ConvLSTM}: An Ultra-Lightweight Efficient Model for Human Activity Recognition on Resource Constrained Devices}

\author{
  \IEEEauthorblockN{Mridankan Mandal}
  \IEEEauthorblockA{Department of Information Technology\\
  Indian Institute of Information Technology, Allahabad\\
  Prayagraj, India\\
  mridankanmandal2006@gmail.com}
}

\maketitle

\begin{abstract}
Human Activity Recognition (HAR) on resource constrained wearables requires models that balance accuracy against strict memory and computational budgets. State of the art lightweight architectures such as TinierHAR (34K parameters) and TinyHAR (55K parameters) achieve strong accuracy, but exceed memory budgets of microcontrollers with limited SRAM once operating system overhead is considered.

We present \modelname{}, an ultra-lightweight convolutional recurrent architecture achieving 11.4K parameters on average through two stage convolutional feature extraction with 4$\times$ temporal pooling, and a single bidirectional LSTM layer. This represents 2.9$\times$ parameter reduction versus TinierHAR and 11.9$\times$ versus DeepConvLSTM while preserving linear $O(N)$ complexity.

Evaluation across eight diverse HAR benchmarks shows that \modelname{} maintains competitive performance within the ultra-lightweight regime: 93.41\% macro F1 on UCI-HAR, 94.46\% on SKODA assembly gestures, and 88.98\% on Daphnet gait freeze detection. Systematic ablation reveals task dependent component contributions where bidirectionality benefits episodic event detection, but provides marginal gains on periodic locomotion. On-device deployment on the Raspberry Pi Pico~2 and ESP32 validates hardware viability under both INT8 quantized and FP32 full-precision paths. Under INT8 quantization, \modelname{} is the only architecture achieving full 8/8 dataset coverage on both platforms, with 72.8~ms average latency on Pico~2 and 97.9\% PyTorch parity on ESP32. Under FP32 deployment, it achieves 100.0\% parity on all successful configurations (8/8 Pico~2, 7/8 ESP32), confirming that all INT8 fidelity degradation is a quantization artifact rather than an architectural limitation.
\end{abstract}

\begin{IEEEkeywords}
Edge AI, Human Activity Recognition, Lightweight Neural Networks, Convolutional LSTM, Ubiquitous computing, Bidirectional LSTM, TinyML, Parameter Efficiency
\end{IEEEkeywords}

\section{Introduction}

Human Activity Recognition (HAR) from wearable inertial sensors enables important applications across healthcare, fitness, and ubiquitous computing \cite{bulling2014tutorial, chen2021deep}. Continuous freeze of gait detection for Parkinson's disease patients \cite{bachlin2010wearable}, real time fall detection for elderly care \cite{micucci2017unimib}, and always on fitness tracking \cite{kwapisz2011activity} share a common requirement of deployment on resource constrained wearable devices where battery life and form factor are important.

Deep learning has substantially improved HAR accuracy through hierarchical temporal pattern learning \cite{hammerla2016deep, ordonez2016deep}. However, an inherent trade-off exists between model capacity and deployment feasibility. DeepConvLSTM \cite{ordonez2016deep}, the canonical convolutional recurrent architecture, achieves strong accuracy, but requires 136K parameters and 15.5M MACs, rendering it impractical for microcontroller based wearables. Recent lightweight alternatives partially address this gap: TinyHAR \cite{zhou2022tinyhar} reduces to 55K parameters using cross channel attention, but introduces $O(N^2)$ sequence complexity, and TinierHAR \cite{bian2025tinierhar} achieves 34K parameters through depthwise separable convolutions and bidirectional GRUs, representing the current state of the art in lightweight HAR.

Yet a deployment gap remains. Microcontrollers with very limited on-chip SRAM must also reserve memory for the operating system, sensor drivers, and communication stacks, leaving only a small fraction for neural network weights and intermediate activations. In practice, even models as compact as TinierHAR can exceed this usable budget once activation memory during inference is considered. This motivates architectures that push parameter counts much lower while still preserving strong temporal modeling capacity.

We introduce \modelname{}, engineered to maximize information density within an ultra-constrained parameter budget. Our contributions:
\begin{itemize}
    \item An 11.4K parameter architecture (averaged across datasets) achieving 2.9$\times$ reduction versus TinierHAR through a two stage convolutional stem with 4$\times$ temporal pooling and single layer bidirectional LSTM, maintaining strict $O(N)$ complexity.
    \item Comprehensive evaluation across eight diverse HAR benchmarks demonstrating competitive accuracy within the ultra-lightweight regime, with performance gaps systematically characterized against dataset complexity.
    \item Systematic ablation studies revealing task dependent component contributions: bidirectionality proves essential for episodic events, but optional for periodic activities.
    \item Cross platform on-device validation on the Raspberry Pi Pico~2 (Arm Cortex-M33) and ESP32 (Xtensa LX6) under both INT8 quantized and FP32 full precision paths, demonstrating full 8/8 INT8 dataset deployment with 72.8~ms and 168.5~ms average latency, respectively, and 100.0\% FP32 parity on all successful configurations, serving as the only architecture in this study achieving complete or near complete coverage on both targets under both precision paths.
\end{itemize}

\begin{figure*}[t]
    \centering
    \includegraphics[width=0.85\textwidth]{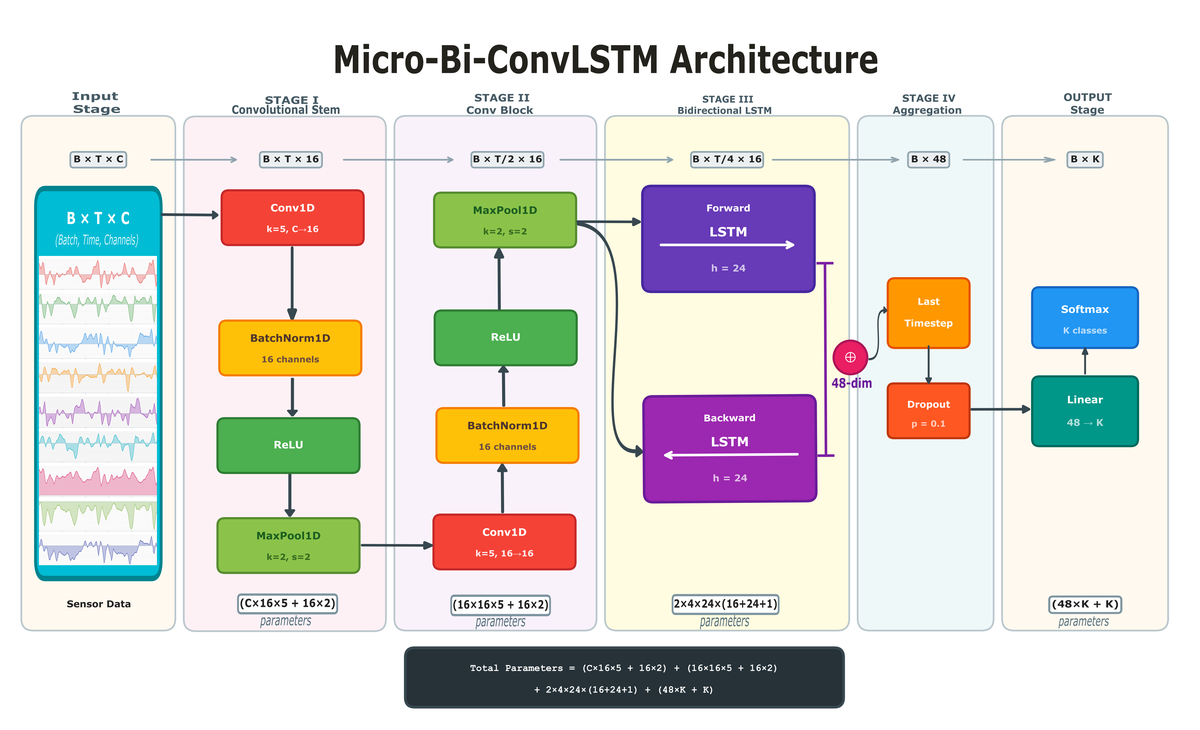}
    \caption{\modelname{} architecture overview.}
    \label{fig:architecture}
\end{figure*}

Figure~\ref{fig:architecture} illustrates the architecture. The design achieves 11.4K parameters on average with strict $O(N)$ complexity, enabling deployment within memory constrained environments.

\section{Related Work}

We review HAR architectures along the accuracy-efficiency spectrum, positioning \modelname{} within the ultra-lightweight regime.

\subsection{Deep Learning Architectures for HAR}

DeepConvLSTM \cite{ordonez2016deep} established the convolutional recurrent paradigm: four convolutional layers (64 filters each) followed by two stacked LSTMs, totaling 136K parameters. While achieving strong accuracy, this architecture proves unsuitable for microcontroller deployment. Subsequent work explored ensemble approaches \cite{guan2017ensembles} and multi-modal fusion \cite{ravi2017deep}, further increasing computational requirements.

Attention based approaches demonstrate superior long range dependency modeling. Abedin et al.\ \cite{abedin2021attend} achieve SOTA performance through attend and discriminate mechanisms, while GlobalFusion \cite{liu2020globalfusion} and AttnSense \cite{ma2019attnsense} leverage multi-level attention for sensor fusion. However, these methods require $>$100K parameters, exceeding edge deployment budgets.

\subsection{Lightweight HAR Architectures}

Recognizing deployment constraints, recent work explores model compression:

\textbf{Efficient Convolutions}: TinierHAR \cite{bian2025tinierhar} achieves 34K parameters through depthwise separable convolutions and bidirectional GRUs, representing current SOTA efficiency. MLP-HAR \cite{zhou2024mlphar} uses purely fully connected layers with FFT preprocessing, achieving 55K parameters, but requiring spectral transformation.

\textbf{Attention-based Compression}: TinyHAR \cite{zhou2022tinyhar} reduces to 55K parameters using cross channel attention, but introduces $O(N^2)$ sequence complexity. Murahari and Pl\"otz \cite{murahari2018attention} analyze attention mechanisms for HAR, finding diminishing returns on shorter sequences typical of wearable data.

\textbf{Knowledge Distillation}: Deng et al.\ \cite{deng2024lhar} transfer knowledge from large teachers to compact students, achieving competitive performance with 50\% parameter reduction, but still exceeding microcontroller budgets.

\subsection{Transformers versus Recurrent Architectures}

The success of Transformers in NLP \cite{vaswani2017attention} has motivated their application to HAR. However, recent studies reveal important trade-offs for wearable deployment. Lattanzi et al.\ \cite{lattanzi2024transformers} systematically evaluate Transformers on tiny devices, finding that while attention mechanisms excel at capturing long range dependencies, their quadratic complexity and memory requirements often exceed microcontroller budgets. Murahari and Pl\"otz \cite{murahari2018attention} demonstrate diminishing returns of attention on shorter sequences typical of HAR (128-256 timesteps), where local temporal patterns dominate. Bock et al.\ \cite{bock2021shallow} show that shallow LSTMs can match deeper architectures on HAR tasks, suggesting that inductive biases of recurrent models align well with the sequential nature of human motion. These findings motivate our choice of BiLSTM over attention based alternatives.

\subsection{Model Compression Techniques}

Post-training quantization \cite{jacob2018quantization, krishnamoorthi2018quantizing} enables INT8 deployment with minimal accuracy loss. Yi et al.\ \cite{yi2023har} combine pruning and quantization for wearable HAR, achieving reduced footprints while maintaining accuracy. Neural Architecture Search \cite{lim2023nas} automates architecture exploration, but requires extensive computational resources.

\subsection{Positioning of \texorpdfstring{\modelname{}}{uBi-ConvLSTM}}

Despite these advances, a gap persists for sub-15K parameter models compatible with the most constrained microcontrollers. Table~\ref{tab:architecture_comparison} compares architectural choices. \modelname{} adheres to the proven convolutional recurrent paradigm while achieving 4$\times$ temporal compression through pooling, substantially reducing LSTM sequence length. Unlike attention based approaches including Transformers, strict $O(N)$ complexity is maintained, with the recurrent inductive bias that aligns with sequential motion data. Unlike depthwise separable convolutions, standard convolutions preserve cross channel information flow important for low dimensional sensor data (3-9 channels typical in smartphone HAR).

\begin{table}[t]
\centering
\caption{Architectural Comparison of HAR Models}
\label{tab:architecture_comparison}
\resizebox{\columnwidth}{!}{%
\begin{tabular}{@{}lccccc@{}}
\toprule
\textbf{Model} & \textbf{Conv Type} & \textbf{Temporal} & \textbf{Attention} & \textbf{Params} ($\downarrow$) & \textbf{Complexity} ($\downarrow$) \\
\midrule
DeepConvLSTM & Standard & 2$\times$LSTM & No & 136K & $O(N)$ \\
TinyHAR & Standard & LSTM+Attn & Yes & 55K & $O(N^2)$ \\
TinierHAR & Depthwise & BiGRU+Attn & Yes & 34K & $O(N^2)$ \\
\modelnameplain{} & Standard & BiLSTM & No & \textbf{11.4K} & $O(N)$ \\
\bottomrule
\end{tabular}%
}\vspace{-2mm}
\end{table}

\section{Methodology}

\subsection{Design Principles}

Based on analysis of HAR data characteristics and edge deployment constraints, we developed the following design principles:

\textbf{P1: Aggressive Temporal Reduction.} Unlike image data with comparable spatial dimensions, HAR exhibits extreme aspect ratios ($T$=128-256 timesteps, $C$=3-9 channels). Long sequences pose challenges for recurrent models because memory scales linearly with sequence length, and gradient propagation degrades over extended horizons. Temporal pooling reduces computational cost while preserving salient motion patterns.

\textbf{P2: Standard Convolutions for Low Dimensional Input.} For sensor data with few channels (3-9 typical), standard convolutions provide superior cross channel feature extraction versus depthwise separable variants. Depthwise operations limit cross channel information flow, important when accelerometer and gyroscope channels contain complementary motion cues.

\textbf{P3: Single Wide Recurrent Layer.} Given parameter constraints, we allocate capacity to a single wide BiLSTM ($H$=24) rather than stacked narrow layers. This maximizes temporal modeling capacity while avoiding redundant hierarchical processing.

\textbf{P4: Bidirectional Processing.} Activities like freeze of gait show temporal asymmetry where onset differs from recovery. Bidirectional LSTMs capture both forward progression and backward context, though ablation studies (Section~\ref{sec:ablation}) reveal this benefit is task dependent.

\textbf{P5: Linear Complexity.} Real time applications require predictable inference time. We maintain strict $O(N)$ complexity, avoiding attention mechanisms that introduce quadratic scaling.

\subsection{Architecture Overview}

\modelname{} comprises five stages optimized for minimal overhead while preserving representational capacity, as illustrated in Figure~\ref{fig:architecture}. Input sensor signals ($C$ channels $\times$ $T$ timesteps) pass through two convolutional blocks with batch normalization, ReLU activation, and 2$\times$ max pooling each, achieving 4$\times$ total temporal compression. A single bidirectional LSTM (hidden dimension 24) processes the compressed sequence, with the final timestep representation feeding the classification head. Parameter count varies with input channels and output classes.

\subsection{Convolutional Stages}

The input $\mathbf{X} \in \mathbb{R}^{B \times T \times C}$ (batch size $B$, sequence length $T$, channels $C$) undergoes two convolutional blocks extracting local temporal features:
\begin{equation}
\mathbf{H}_1 = \text{ReLU}\left(\text{BN}\left(\mathbf{W}_1 * \mathbf{X} + \mathbf{b}_1\right)\right)
\end{equation}
where $\mathbf{W}_1 \in \mathbb{R}^{F \times C \times K}$ with $F=16$ filters and kernel size $K=5$. Max pooling after each block provides 2$\times$ temporal compression, yielding total 4$\times$ reduction: $T \rightarrow T/4$. The receptive field spans 9 timesteps, which is sufficient for activity relevant patterns like gait cycles and gesture transitions, computed as $R = 1 + \sum_{i=1}^{L} (K_i - 1) \cdot \prod_{j=1}^{i-1} S_j$.

\subsection{Bidirectional LSTM}

Compressed features pass through a single layer Bidirectional LSTM with hidden dimension $H=24$:
\begin{align}
\mathbf{i}_t &= \sigma\left(\mathbf{W}_{ii}\mathbf{x}_t + \mathbf{W}_{hi}\mathbf{h}_{t-1} + \mathbf{b}_i\right) \\
\mathbf{f}_t &= \sigma\left(\mathbf{W}_{if}\mathbf{x}_t + \mathbf{W}_{hf}\mathbf{h}_{t-1} + \mathbf{b}_f\right) \\
\mathbf{c}_t &= \mathbf{f}_t \odot \mathbf{c}_{t-1} + \mathbf{i}_t \odot \tanh(\mathbf{W}_{ig}\mathbf{x}_t + \mathbf{W}_{hg}\mathbf{h}_{t-1}) \\
\mathbf{h}_t &= \sigma(\mathbf{W}_{io}\mathbf{x}_t + \mathbf{W}_{ho}\mathbf{h}_{t-1} + \mathbf{b}_o) \odot \tanh(\mathbf{c}_t)
\end{align}
Bidirectional processing concatenates forward and backward passes into a 48 dimensional representation. We use last timestep aggregation: $\mathbf{h}_{\text{agg}} = \mathbf{H}[:, -1, :]$.

\textbf{Design Rationale}: BiLSTM was selected over alternatives for three reasons. (1)~Bidirectionality captures temporal asymmetry in episodic events where onset and recovery patterns differ. (2)~LSTM's separate cell state provides superior gradient flow versus GRU for longer compressed sequences ($T/4$ = 32 timesteps). (3)~Last timestep aggregation avoids attention overhead while the final bidirectional state synthesizes both full forward sequence context and complete backward information.

\subsection{Classification Head}

The aggregated representation passes through dropout regularization and a linear classifier:
\begin{equation}
\hat{\mathbf{y}} = \text{Softmax}\left(\mathbf{W}_{\text{out}} \cdot \text{Dropout}(\mathbf{h}_{\text{agg}}) + \mathbf{b}_{\text{out}}\right)
\end{equation}

\subsection{Parameter and Complexity Analysis}

The architecture's parameter count scales with input channels and output classes, ranging from approximately 10K to 32K across datasets (Table~\ref{tab:efficiency}). The BiLSTM layer dominates at roughly 75\% of total parameters, computed as $P_{\text{LSTM}} = 2 \times 4 \times H \times (F + H + 1)$ where $H$ is hidden size and $F$ is input features. Total MACs scale linearly with sequence length $T$ and input channels, with Table~\ref{tab:macs_per_dataset} showing 245K--1.14M MACs across configurations. This represents 32--42$\times$ reduction versus DeepConvLSTM and 5--42$\times$ versus TinyHAR, while maintaining strict $O(N)$ complexity.

\section{Experimental Setup}

\subsection{Datasets}

Evaluation was done on eight publicly available HAR benchmarks spanning diverse sensing modalities, subject populations, and activity types (Table~\ref{tab:datasets}). UCI-HAR \cite{anguita2013public} provides smartphone accelerometer and gyroscope data from 30 subjects performing six locomotion activities (walking, climbing stairs, sitting, standing, lying) at 50~Hz. MotionSense \cite{malekzadeh2019mobile} similarly captures smartphone IMU data from 24 subjects, but includes device attitude information. WISDM \cite{kwapisz2011activity} offers accelerometer only data from 36 subjects at lower 20~Hz sampling, testing model robustness to reduced input dimensionality.

PAMAP2 \cite{reiss2012introducing} and Opportunity \cite{chavarriaga2013opportunity} represent more challenging multi-sensor configurations. PAMAP2 fuses data from three body worn IMUs (19 channels) across 12 activities from 9 subjects, while Opportunity provides 79 channels from a dense sensor network with substantial class imbalance. UniMiB-SHAR \cite{micucci2017unimib} focuses on fall detection with 9 activity classes. SKODA \cite{stiefmeier2008wearable} targets industrial assembly gesture recognition with 30 channel sensor gloves at 98~Hz from a single subject performing repetitive manufacturing tasks. Daphnet \cite{bachlin2010wearable} addresses Parkinson's disease gait freeze detection, a binary classification task with severe class imbalance (approximately 10\% freeze events) requiring high sensitivity for clinical utility.

\begin{table}[t]
\centering
\caption{Dataset Characteristics}
\label{tab:datasets}
\resizebox{\columnwidth}{!}{%
\begin{tabular}{@{}lccccccc@{}}
\toprule
\textbf{Dataset} & \textbf{Type} & \textbf{Subjects} & \textbf{Classes} & \textbf{Channels} & \textbf{Frequency} & \textbf{Window} \\
\midrule
UCI-HAR \cite{anguita2013public} & Phone & 30 & 6 & 9 & 50 Hz & 128 \\
MotionSense \cite{malekzadeh2019mobile} & Phone & 24 & 6 & 6 & 50 Hz & 128 \\
WISDM \cite{kwapisz2011activity} & Phone & 36 & 6 & 3 & 20 Hz & 128 \\
PAMAP2 \cite{reiss2012introducing} & Body IMU & 9 & 12 & 19 & 100 Hz & 128 \\
Opportunity \cite{chavarriaga2013opportunity} & Body IMU & 4 & 5 & 79 & 30 Hz & 128 \\
UniMiB \cite{micucci2017unimib} & Phone & 30 & 9 & 3 & 50 Hz & 128 \\
SKODA \cite{stiefmeier2008wearable} & Glove IMU & 1 & 11 & 30 & 98 Hz & 98 \\
Daphnet \cite{bachlin2010wearable} & Body IMU & 10 & 2 & 9 & 64 Hz & 64 \\
\bottomrule
\end{tabular}%
}
\end{table}

\subsection{Preprocessing and Signal Conditioning}

Raw sensor signals undergo dataset specific preprocessing to address noise characteristics. For datasets with known high frequency noise (industrial and clinical settings), 4th order Butterworth low pass filtering was applied to: PAMAP2 (10~Hz cutoff), SKODA (5~Hz cutoff for motor vibration), and Daphnet (12~Hz cutoff). All features are z-score normalized per channel using training set statistics. Sliding windows with 50\% overlap generate samples.

For multi-subject datasets (UCI-HAR, MotionSense, WISDM, PAMAP2, UniMiB, Daphnet), subject wise cross validation splits were used, ensuring no data leakage between training and evaluation, that is, samples from the same subject never appear in both training and test sets. For single subject datasets (SKODA) and Opportunity with predefined protocol splits, the standard train/test partitions from the original publications were used. This subject independent evaluation protocol provides unbiased generalization estimates for deployment on unseen users.

\subsection{Training Protocol}

All models train for a maximum of 200 epochs using AdamW \cite{loshchilov2019decoupled} with cosine annealing learning rate scheduling. Early stopping with patience of 10 epochs was used, based on validation F1-score. For imbalanced datasets (Daphnet, Opportunity, PAMAP2), inverse frequency class weighting is applied in the cross-entropy loss.

Hyperparameter optimization (HPO) uses Optuna \cite{akiba2019optuna} with 50 Tree-structured Parzen Estimator (TPE) trials per dataset. Each HPO trial trains for 10 epochs with early stopping patience of 5 epochs to enable rapid exploration. The search space covers learning rate ($10^{-4}$ to $10^{-2}$, log scale), weight decay ($10^{-5}$ to $5 \times 10^{-2}$, log scale), and dropout (0.0 to 0.5, uniform). Architecture parameters remain frozen during HPO to isolate training dynamics from structural effects.

Final training uses the optimized hyperparameters for 200 epochs with early stopping patience of 10 epochs. We report macro F1-score mean and standard deviation across 5 random seeds generated from a master seed for reproducibility.

\subsection{Baseline Models}

Comparison was made against three representative architectures spanning the efficiency spectrum, all trained with identical HPO and training protocols for fair comparison.

\textbf{DeepConvLSTM} \cite{ordonez2016deep} serves as the canonical high capacity baseline following the original specification: four convolutional layers with 64 filters each (kernel size 5, same padding), followed by two stacked unidirectional LSTM layers with 64 hidden units per layer. This configuration totals approximately 136K parameters.

\textbf{TinyHAR} \cite{zhou2022tinyhar} represents modern attention based lightweight design.  The configuration from the original implementation was used: 24 filters across 4 convolutional layers with kernel size 5, and cross channel self attention for temporal fusion. This achieves approximately 55K parameters.

\textbf{TinierHAR} \cite{bian2025tinierhar} provides the most aggressive prior compression using depthwise separable convolutions and bidirectional GRUs. The model is configured with 8 filters across 4 depthwise separable convolutional blocks with 16 GRU hidden units per direction, totaling approximately 34K parameters.

\section{Results}

\subsection{Predictive Performance}

Table~\ref{tab:f1_comparison} presents F1 scores across all benchmarks. Several patterns emerge from these results.

\textbf{Competitive Performance in Ultra-Lightweight Regime}: Despite 2.9$\times$ fewer parameters than TinierHAR, \modelname{} achieves comparable accuracy on smartphone IMU datasets. On UCI-HAR, the gap compared to TinyHAR (the best performer) is 3.12\%, while on MotionSense the gap narrows to 1.02\%.

\textbf{Strong Medical Monitoring Performance}: On Daphnet, gait freeze detection, \modelname{} achieves 88.98\% F1, competitive with TinierHAR's 89.84\%. Both substantially outperform the attention based TinyHAR (86.42\%), suggesting that recurrent architectures better capture the temporal dynamics of episodic freezing events than attention mechanisms for this binary classification task.

\begin{table}[t]
\centering
\caption{F1 Score Comparison (Macro \%) ($\uparrow$)}
\label{tab:f1_comparison}
\resizebox{\columnwidth}{!}{%
\begin{tabular}{@{}lcccc@{}}
\toprule
\textbf{Dataset} & \textbf{\modelnameplain{}} & \textbf{DeepConvLSTM} & \textbf{TinyHAR} & \textbf{TinierHAR} \\
\midrule
UCI-HAR & 93.41$\pm$0.35 & 93.61$\pm$0.56 & 96.53$\pm$0.41 & 96.37$\pm$0.59 \\
MotionSense & 91.65$\pm$0.43 & 91.64$\pm$0.75 & 92.67$\pm$0.67 & 91.99$\pm$0.60 \\
WISDM & 73.17$\pm$12.4 & 81.25$\pm$2.55 & 77.09$\pm$4.95 & 83.06$\pm$3.24 \\
PAMAP2 & 60.75$\pm$1.76 & 66.20$\pm$2.72 & 73.22$\pm$3.58 & 74.07$\pm$1.16 \\
Opportunity & 87.58$\pm$0.73 & 87.21$\pm$0.63 & 88.69$\pm$0.38 & 87.09$\pm$0.90 \\
UniMiB & 79.43$\pm$1.66 & 84.12$\pm$2.23 & 77.61$\pm$2.23 & 79.67$\pm$4.45 \\
SKODA & 94.46$\pm$1.31 & 95.25$\pm$1.03 & 97.01$\pm$0.53 & 96.99$\pm$0.76 \\
Daphnet & 88.98$\pm$1.64 & 88.19$\pm$1.89 & 86.42$\pm$3.64 & 89.84$\pm$1.90 \\
\midrule
\textbf{Average} & 83.68 & 85.93 & 86.16 & \textbf{87.39} \\
\bottomrule
\end{tabular}%
}
\end{table}

\begin{figure*}[t]
    \centering
    \includegraphics[width=0.9\textwidth]{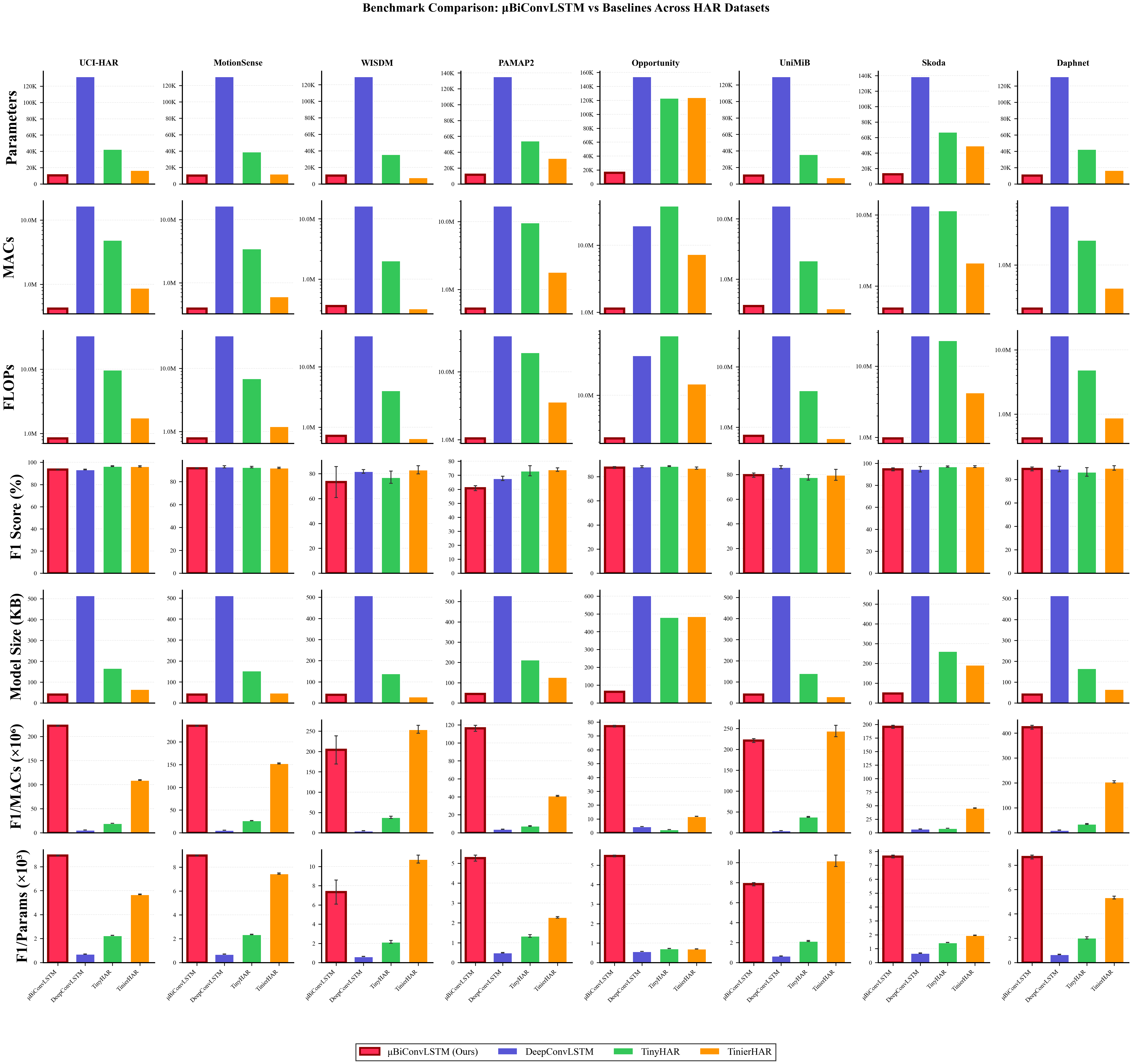}
    \caption{Parameters, MACs, FLOPs, Model Size (in KB), F1-score per million MACs, and F1-score per thousand parameters distributions across architectures and datasets. Box plots show mean, and standard deviations across five random seeds.}
    \label{fig:benchmark_comparison}
\end{figure*}

\textbf{Class Imbalance Sensitivity}: The performance gap between \modelname{} and larger baselines varies systematically with dataset characteristics. PAMAP2 shows a 13.3\% F1-score gap as compared to TinierHAR due to extreme class imbalance where rare activities (rope jumping, cycling) constitute $<$10\% of samples. Ultra-lightweight models prioritize majority classes under capacity constraints, degrading macro F1-score. In contrast, Opportunity's balanced gesture distribution enables competitive performance (87.58\% vs.\ 87.09\%) despite its 79 channel density, demonstrating that class distribution impacts lightweight models more than sensor dimensionality. The WISDM results show high variance (12.4\% std), likely due to the dataset's lower sampling rate (20~Hz) and minimal channel count (3) creating challenging optimization dynamics.

Figure~\ref{fig:benchmark_comparison} visualizes performance distributions. \modelname{} (leftmost in each group) maintains competitive variance despite 2.9$\times$ fewer parameters than TinierHAR, particularly on UCI-HAR and SKODA, both well structured datasets with clear activity boundaries.

\begin{figure}[t]
    \centering
    \includegraphics[width=\columnwidth]{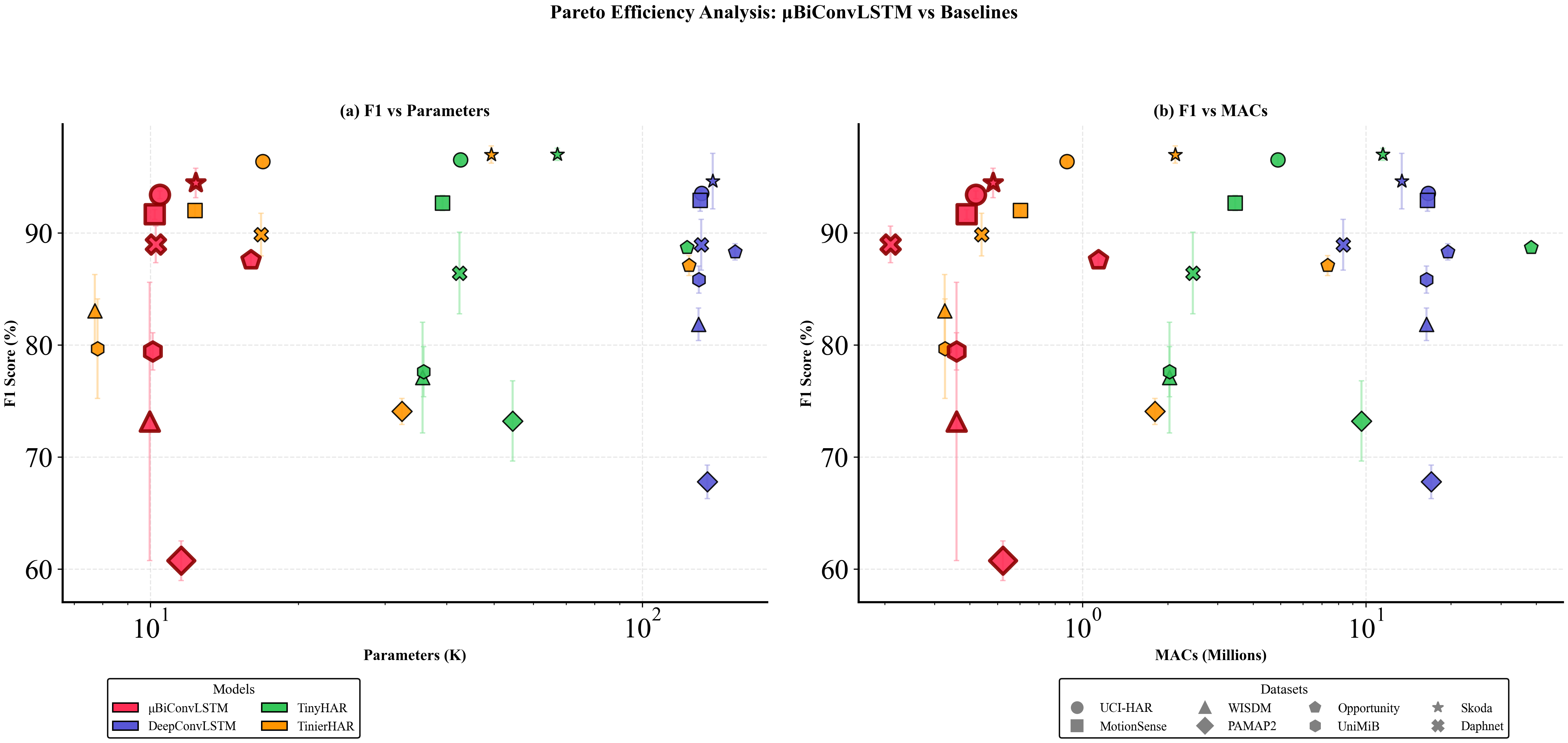}
    \caption{Pareto frontier: F1-score versus MACs (log scale). Each point represents one model-dataset combination.}
    \label{fig:pareto}
\end{figure}

\begin{figure}[t]
    \centering
    \includegraphics[width=\columnwidth]{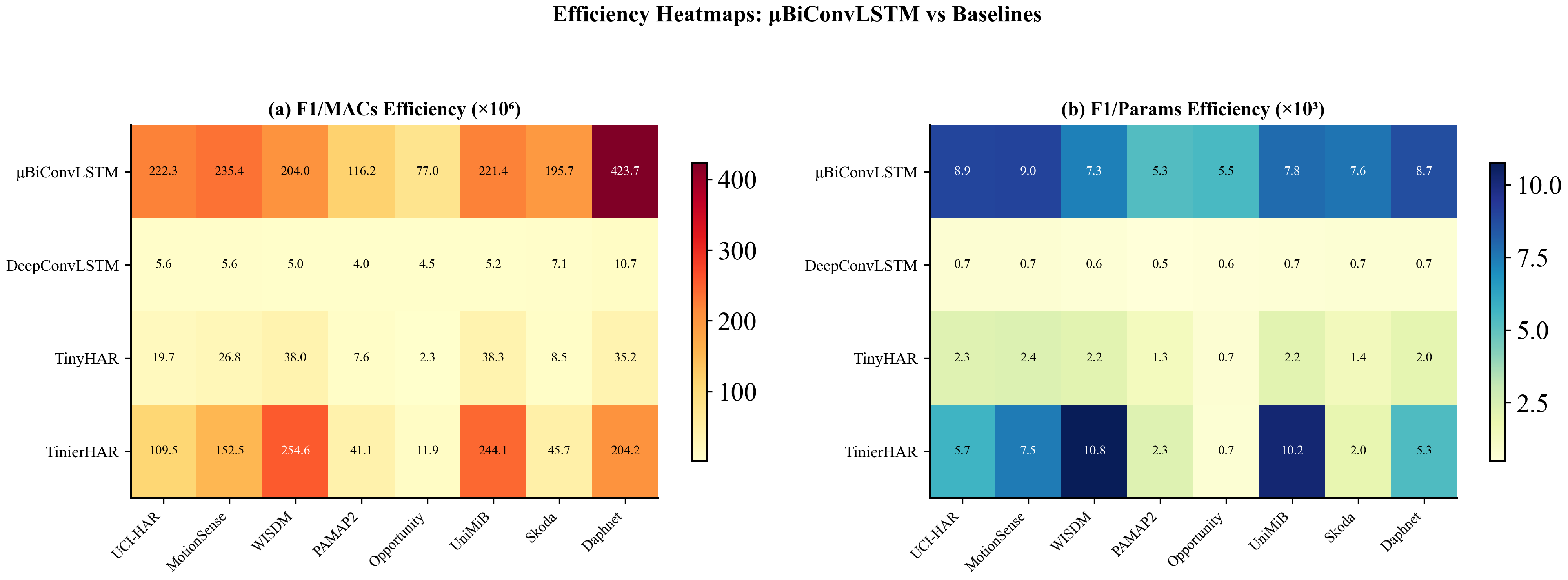}
    \caption{Efficiency heatmap: F1-score per thousand parameters across datasets. Darker cells indicate higher parameter efficiency.}
    \label{fig:efficiency_heatmap}
\end{figure}

\subsection{Computational Efficiency}

Table~\ref{tab:efficiency} quantifies efficiency advantages. \modelname{} achieves 7.34 F1 per thousand parameters and 172.5 F1 per million MACs. Compared to TinierHAR, this represents 2.8$\times$ and 3.4$\times$ improvements respectively. This efficiency stems from our architectural choices: temporal pooling reduces LSTM sequence length by 4$\times$, and the single layer bidirectional design eliminates redundant stacked processing.

\begin{table}[t]
\centering
\caption{Computational Efficiency Comparison}
\label{tab:efficiency}
\resizebox{\columnwidth}{!}{%
\begin{tabular}{@{}lcccc@{}}
\toprule
\textbf{Metric} & \textbf{\modelnameplain{}} & \textbf{DeepConvLSTM} & \textbf{TinyHAR} & \textbf{TinierHAR} \\
\midrule
Parameters ($\downarrow$) & \textbf{11.4K} & 135.6K & 55.1K & 33.5K \\
MACs ($\downarrow$) & \textbf{485K} & 15.51M & 9.29M & 1.73M \\
F1/K-Params ($\uparrow$) & \textbf{7.34} & 0.63 & 1.56 & 2.61 \\
F1/M-MACs ($\uparrow$) & \textbf{172.5} & 5.54 & 9.27 & 50.5 \\
Complexity ($\downarrow$) & $O(N)$ & $O(N)$ & $O(N^2)$ & $O(N^2)$ \\
\bottomrule
\end{tabular}%
}
\end{table}

Figure~\ref{fig:pareto} reveals that \modelname{} occupies the Pareto-optimal region for the ultra-lightweight regime (top left), achieving competitive accuracy at substantially lower computational cost than all baselines. The separation from TinierHAR demonstrates that further parameter reduction remains viable without catastrophic accuracy loss.

The efficiency heatmap (Figure~\ref{fig:efficiency_heatmap}) confirms that \modelname{} parameter efficiency advantage is consistent across all benchmarks rather than dataset specific, indicating a fundamental architectural benefit. Table~\ref{tab:macs_per_dataset} details per dataset MACs. Notably, even on 79 channel Opportunity, \modelname{} requires only 1.14M MACs compared to TinyHAR's 47.8M, a 42$\times$ reduction attributable to avoiding quadratic attention complexity.

\begin{table}[t]
\centering
\caption{MACs by Dataset (thousands) ($\downarrow$)}
\label{tab:macs_per_dataset}
\resizebox{\columnwidth}{!}{%
\begin{tabular}{@{}lrrrr@{}}
\toprule
\textbf{Dataset} & \textbf{\modelnameplain{}} & \textbf{DeepConvLSTM} & \textbf{TinyHAR} & \textbf{TinierHAR} \\
\midrule
UCI-HAR & \textbf{420} & 16,620 & 6,050 & 1,080 \\
MotionSense & \textbf{389} & 16,490 & 4,270 & 740 \\
WISDM & \textbf{359} & 16,370 & 2,480 & 402 \\
PAMAP2 & \textbf{523} & 17,030 & 12,010 & 2,210 \\
Opportunity & \textbf{1,140} & 19,480 & 47,770 & 8,970 \\
UniMiB & \textbf{359} & 16,370 & 2,480 & 402 \\
SKODA & \textbf{444} & 13,290 & 12,360 & 1,910 \\
Daphnet & \textbf{245} & 8,110 & 2,640 & 399 \\
\midrule
\textbf{Average} & \textbf{485} & 15,510 & 9,290 & 1,730 \\
\bottomrule
\end{tabular}%
}
\end{table}

\subsection{Desktop Profiling}

Table~\ref{tab:quantization} presents INT8 post training quantization results. \modelname{} shows remarkable quantization robustness with only 0.21\% average F1 degradation. Interestingly, PAMAP2 and Opportunity show slight \textit{improvements} after quantization ($-$0.06\%), suggesting that reduced numerical precision provides implicit regularization beneficial for these noisy multi sensor datasets. The Daphnet degradation (1.09\%) is acceptable given the binary classification task's inherent robustness to small prediction shifts.

\begin{table}[t]
\centering
\caption{INT8 Quantization Impact on \modelnameplain{}}
\label{tab:quantization}
\begin{tabular}{@{}lccc@{}}
\toprule
\textbf{Dataset} & \textbf{FP32 F1} ($\uparrow$) & \textbf{INT8 F1} ($\uparrow$) & \textbf{Degradation} ($\downarrow$) \\
\midrule
UCI-HAR & 0.9269 & 0.9269 & \textbf{0.00\%} \\
MotionSense & 0.9088 & 0.9046 & 0.42\% \\
WISDM & 0.7684 & 0.7682 & 0.02\% \\
PAMAP2 & 0.6373 & 0.6379 & -0.06\% \\
Opportunity & 0.8703 & 0.8709 & -0.06\% \\
SKODA & 0.9571 & 0.9564 & 0.07\% \\
Daphnet & 0.8813 & 0.8705 & 1.09\% \\
\midrule
\textbf{Average} & 0.8500 & 0.8479 & \textbf{0.21\%} \\
\bottomrule
\end{tabular}
\end{table}

\begin{figure}[t]
    \centering
    \includegraphics[width=\columnwidth]{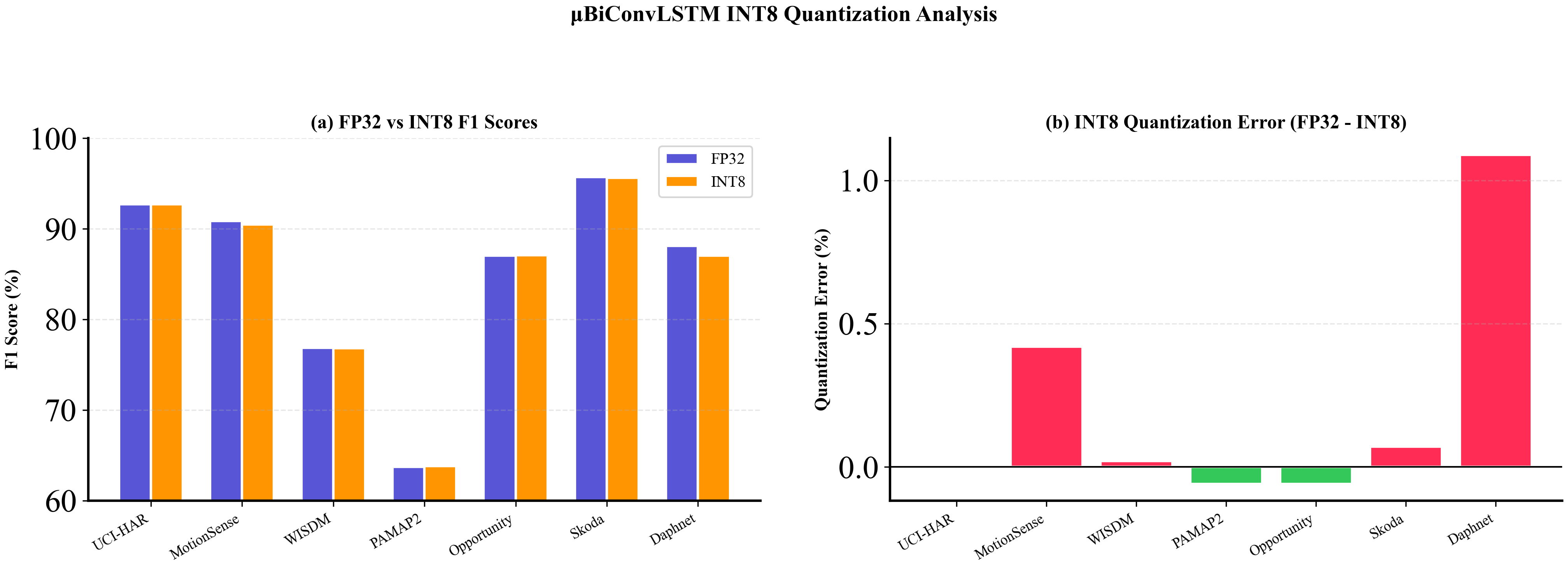}
    \caption{FP32 versus INT8 F1-scores for \modelnameplain{} across datasets.}
    \label{fig:desktop_quantization}
\end{figure}

This quantization resilience likely stems from our architectural simplicity: batch normalization after each convolution constrains activation ranges, the single LSTM layer avoids error accumulation through stacked quantized operations, and the 4$\times$ temporal pooling reduces the number of quantized computations. Figure~\ref{fig:desktop_quantization} visualizes this robustness, where the near diagonal alignment demonstrates quantization robustness, with average degradation of only 0.21\%.

Table~\ref{tab:memory} presents INT8 memory requirements across all datasets, with overhead expressed as percentage increase relative to \modelname{}. Our architecture averages 23.0~KB, while TinierHAR requires 66\% more memory, TinyHAR requires 385\% more, and DeepConvLSTM requires 1,056\% more. This positions \modelname{} as the architecture consistently viable for memory constrained edge devices across all tested configurations.

\begin{table}[t]
\centering
\caption{Memory Footprint (INT8, KB) ($\downarrow$) with Overhead vs \modelnameplain{}}
\label{tab:memory}
\resizebox{\columnwidth}{!}{%
\begin{tabular}{@{}lrrrr@{}}
\toprule
\textbf{Dataset} & \textbf{\modelnameplain{}} & \textbf{TinierHAR} & \textbf{TinyHAR} & \textbf{DeepConvLSTM} \\
\midrule
UCI-HAR & \textbf{21.2} & 34.1 & 84.7 & 264.3 \\
MotionSense & \textbf{20.7} & 33.6 & 72.2 & 263.6 \\
WISDM & \textbf{20.2} & 33.0 & 59.7 & 262.8 \\
PAMAP2 & \textbf{23.4} & 37.7 & 124.7 & 267.8 \\
Opportunity & \textbf{32.4} & 56.7 & 305.0 & 295.5 \\
UniMiB & \textbf{20.5} & 33.3 & 59.7 & 263.1 \\
SKODA & \textbf{25.1} & 43.1 & 128.4 & 256.3 \\
Daphnet & \textbf{20.8} & 33.6 & 58.3 & 252.6 \\
\midrule
\textbf{Average} & \textbf{23.0} & 38.1 & 111.6 & 265.8 \\
\textbf{Overhead vs \modelnameplain{}} & -- & +66\% & +385\% & +1,056\% \\
\bottomrule
\end{tabular}%
}
\end{table}

\modelname{}'s footprint scales gracefully with input dimensionality: the 79 channel Opportunity dataset requires only 32.4~KB (41\% increase from the 3 channel minimum), whereas TinyHAR increases to 305~KB (411\% increase) due to attention's quadratic channel scaling. This characteristic makes \modelname{} particularly suitable for multi-sensor wearable platforms where memory budgets are fixed, but sensor configurations may vary.

\subsection{INT8 Quantized Deployment: Raspberry Pi Pico 2}

To validate deployment feasibility beyond desktop profiling, INT8 quantized TensorFlow Lite Micro models are deployed, on the Raspberry Pi Pico~2 (RP2350, dual-core Arm Cortex-M33, 520~KB SRAM, 4~MB flash). All models are compiled with the standard \texttt{pico-tflmicro} runtime, and executed on the device with identical arena configurations. We report tensor arena usage, model binary size, inference latency, and PyTorch parity, which is the percentage of test samples where the on-device prediction agrees with the original FP32 PyTorch model's prediction.

Table~\ref{tab:pico2_deployment} presents family level deployment results. \modelname{} is the only architecture that completes all 8/8 datasets on this platform, achieving the lowest average latency (72.8~ms), the smallest average model binary (275~KB), and the highest average parity (85.7\%). TinierHAR fits in memory for all datasets but shows lower fidelity (54.2\% average parity), and 3.1$\times$ higher latency. TinyHAR fails on one dataset (Opportunity) due to buffer allocation limits, and several successful runs show severe prediction drift. DeepConvLSTM is the clearest negative result: 7/8 datasets fail during tensor allocation, and the only successful run (Daphnet) degrades to 25.5\% parity, which is effectively random for a binary task.

\begin{table}[t]
\centering
\caption{Pico~2 On-Device Deployment Summary}
\label{tab:pico2_deployment}
\resizebox{\columnwidth}{!}{%
\begin{tabular}{@{}lcccccc@{}}
\toprule
\textbf{Model} & \textbf{Runs} ($\uparrow$) & \textbf{Fails} ($\downarrow$) & \textbf{Latency (ms)} ($\downarrow$) & \textbf{Arena (B)} ($\downarrow$) & \textbf{Model (B)} ($\downarrow$) & \textbf{Parity (\%)} ($\uparrow$) \\
\midrule
\modelnameplain{} & \textbf{8} & \textbf{0} & \textbf{72.8} & \textbf{109,630} & \textbf{275,088} & \textbf{85.7} \\
TinyHAR & 7 & 1 & 354.1 & 123,188 & 195,484 & 60.4 \\
TinierHAR & 8 & 0 & 229.3 & 183,974 & 312,433 & 54.2 \\
DeepConvLSTM & 1 & 7 & 768.8 & 244,084 & 865,304 & 25.5 \\
\bottomrule
\end{tabular}%
}
\end{table}

Per dataset analysis reveals that \modelname{} maintains strong fidelity across heterogeneous benchmarks. On UCI-HAR, it preserves 89.9\% parity at 78.9~ms latency. On Opportunity (79 channels) it reaches 94.7\% parity despite the increased arena footprint (134~KB). The highest parity is achieved on SKODA (95.9\%), where the model's convolutional stem aligns naturally with the repetitive gesture structure. Even on Daphnet, where the binary freeze/no-freeze boundary is sensitive to quantization noise, \modelname{} sustains 72.0\% parity versus DeepConvLSTM's 25.5\%, confirming that architectural compactness translates to more predictable on-device behavior.

The failure modes of the baselines are instructive. DeepConvLSTM's dominant failure is tensor arena allocation exhaustion: it's $\sim$1.9~MB model binaries, and correspondingly large intermediate tensors exceed the Pico~2's contiguous SRAM budget before inference begins. TinyHAR's Opportunity failure is a runtime buffer resize overflow (requested 493~KB against 302~KB available), indicating that attention based intermediate activations scale poorly under fixed arena constraints. These results confirm that the INT8 model size reported in desktop profiling (Table~\ref{tab:efficiency}) is a necessary, but insufficient predictor of on-device viability. Actual deployment must also account for runtime tensor allocation, arena fragmentation, and kernel scratch memory.

\subsection{INT8 Quantized Deployment: ESP32}

Deployment validation is also done on the ESP32-D0WD-V3 (single-core Xtensa LX6, 520~KB SRAM, 4~MB flash, no PSRAM), a widely deployed platform in commercial IoT. Models are deployed through the native ESP-IDF \texttt{esp-tflite-micro} pipeline with optimized arena sizing. TinierHAR required additional memory path tuning: single core mode with IRAM exposed as 8 bit accessible heap, and a 176~KB arena allocation. An initial deployment sweep revealed catastrophic parity collapses in several baseline quantized exports. Investigation traced the root cause to activation range loss in the original INT8 export recipe, rather than to hardware limitations. A corrected mixed quantized export using INT16 activations with INT8 weights was therefore evaluated for the affected baseline rows, and promoted where it materially improved on-device fidelity.

Table~\ref{tab:esp32_deployment} summarizes the corrected results. \modelname{} again achieves full 8/8 dataset coverage, the highest average PyTorch parity (97.9\%), and moderate latency (168.5~ms). On-device parity on ESP32 is considerably higher than on Pico~2 (97.9\% vs.\ 85.7\%), due to the ESP32's wider ALU datapath, and more favorable fixed point rounding behavior in the Xtensa \texttt{esp-nn} kernels. TinierHAR deploys on 6/8 datasets with an improved average parity of 90.8\% after the corrected quantized export, making it a credible secondary result on this platform. TinyHAR deploys on 4/8 datasets with 88.6\% average parity, a notable recovery from the earlier near-zero fidelity figures, though one previously successful dataset (UCI-HAR) became memory-limited after the corrected export because the more faithful quantized graph required a larger internal SRAM allocation. DeepConvLSTM fails on all 8/8 datasets due to tensor arena allocation failure on the no-PSRAM ESP32, confirming that this architecture class is fundamentally incompatible with sub-520~KB SRAM platforms.

\begin{table}[t]
\centering
\caption{ESP32 On-Device Deployment Summary}
\label{tab:esp32_deployment}
\resizebox{\columnwidth}{!}{%
\begin{tabular}{@{}lccccc@{}}
\toprule
\textbf{Model} & \textbf{Runs} ($\uparrow$) & \textbf{Fails} ($\downarrow$) & \textbf{Latency (ms)} ($\downarrow$) & \textbf{Model (B)} ($\downarrow$) & \textbf{Parity (\%)} ($\uparrow$) \\
\midrule
\modelnameplain{} & \textbf{8} & \textbf{0} & \textbf{168.5} & \textbf{275,088} & \textbf{97.9} \\
TinyHAR & 4 & 4 & 487.0 & 218,202 & 88.6 \\
TinierHAR & 6 & 2 & 204.5 & 312,433 & 90.8 \\
DeepConvLSTM & 0 & 8 & --- & 1,758,060 & --- \\
\bottomrule
\end{tabular}%
}
\end{table}

\begin{figure}[t]
    \centering
    \includegraphics[width=\columnwidth]{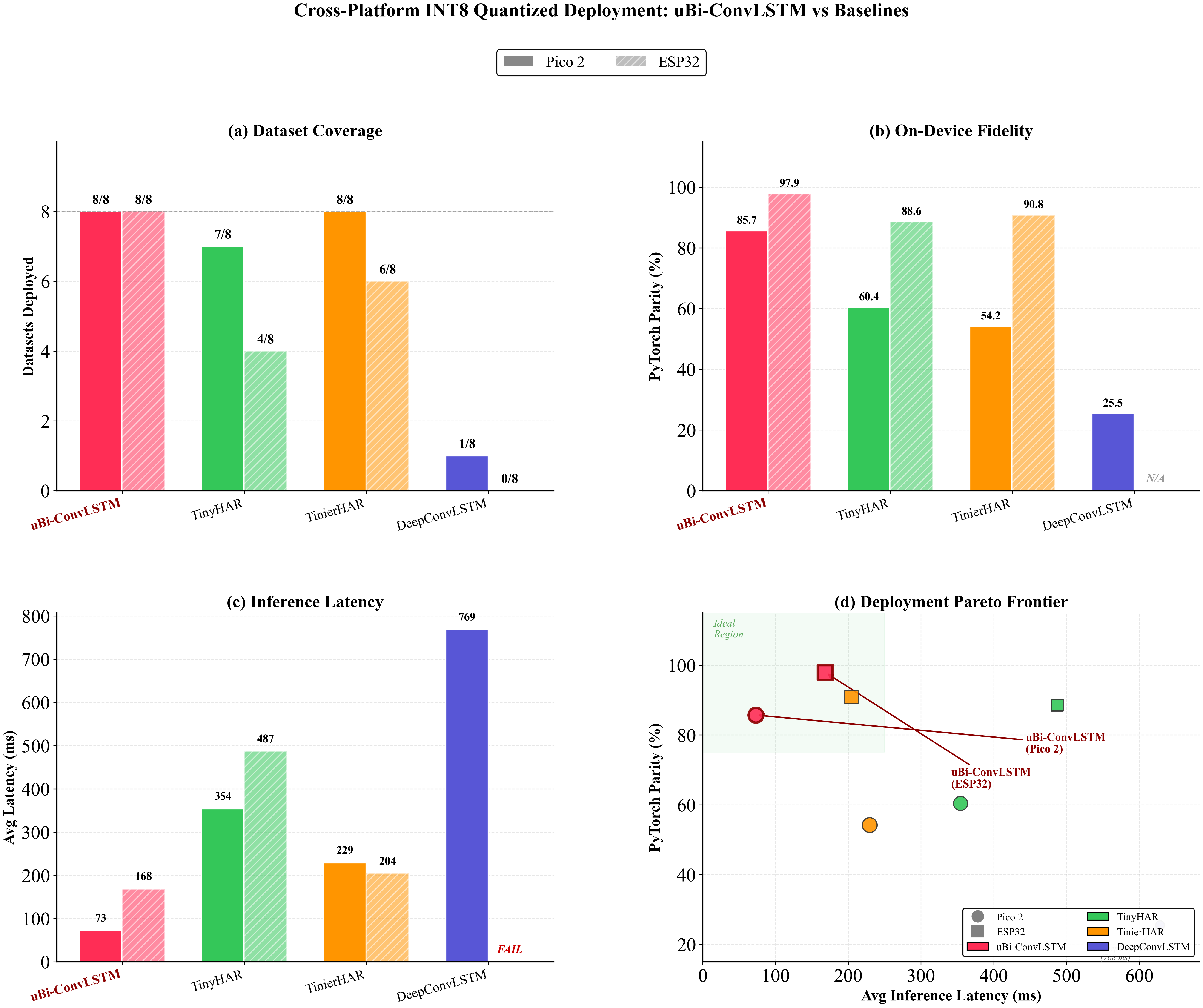}
    \caption{Cross platform INT8 quantized deployment comparison across Pico~2 and ESP32: (a)~dataset coverage out of eight benchmarks, (b)~average PyTorch parity measuring on-device INT8 fidelity, (c)~average inference latency, and (d)~deployment Pareto frontier plotting latency against parity with marker size proportional to dataset coverage. Hatched bars denote ESP32, and solid bars denote Pico~2.}
    \label{fig:deployment_comparison}
\end{figure}

Per dataset ESP32 results reveal additional deployment insights. \modelname{} achieves near perfect parity on several benchmarks: 99.3\% on UCI-HAR, 99.6\% on SKODA, and 99.2\% on Opportunity. These figures exceed the desktop INT8 quantization degradation bounds (Table~\ref{tab:efficiency}), confirming that the quantized model generalizes to heterogeneous hardware backends. The Daphnet configuration achieves the lowest latency (95.2~ms) due to its smaller 144~KB arena, and reduced model binary (155~KB), displaying the favorable latency memory scaling of the architecture.

The corrected baseline results show an important methodological concern. The earlier catastrophic parity collapses observed for TinyHAR and TinierHAR were not purely hardware limitations, but were caused by activation range loss in the standard full INT8 export path. Replacing the export recipe with INT16 activations and INT8 weights recovered meaningful fidelity for TinyHAR on MotionSense (from near-zero to 88.1\%), and for TinierHAR on MotionSense (to 91.6\%). This recovery came at a cost: the more faithful quantized graphs sometimes required larger tensor arenas, pushing previously deployable configurations past the internal SRAM limit. TinyHAR on UCI-HAR is an example, where the corrected export fits on desktop, but fails to reserve a contiguous 256~KB arena on the ESP32.

The ESP32 failure taxonomy remains practically informative. Two distinct failure classes emerge: (1)~tensor arena allocation failures, where the board cannot reserve the requested contiguous internal SRAM block at startup (dominant for DeepConvLSTM, and four TinyHAR datasets), and (2)~\texttt{AllocateTensors} resize failures, where the arena exists but planner scratch memory exceeds the residual budget (remaining TinierHAR failures on PAMAP2 and Opportunity). This distinction matters because the second class indicates the native inference path is valid, and close to working, as it is no longer an ABI or kernel compatibility problem, but a quantitative arena budget shortfall. Figure~\ref{fig:deployment_comparison} visualizes the cross platform deployment landscape, showing how the \modelname{} occupies the ideal top left region of the Pareto space on both platforms, combining the lowest latency with the highest fidelity.

Across both platforms under INT8 quantization, \modelname{} uniquely combines full dataset coverage with low latency, and high inference fidelity. Its 4$\times$ temporal pooling, and single layer BiLSTM, yield smaller tensor arenas, shorter inference paths, and predictable runtime under fixed memory constraints, empirically validating the efficiency claims of Sections~V-A and V-B.

\subsection{FP32 Full-Precision Deployment}

To isolate quantization induced fidelity loss from architectural deployment limitations, FP32 full-precision TFLite models are deployed, on both platforms. The FP32 path eliminates the INT8 rounding error entirely but produces larger model binaries and tensor arenas, reducing dataset coverage on memory constrained MCUs. Tables~\ref{tab:pico2_fp32} and~\ref{tab:esp32_fp32} summarize the FP32 results.

\begin{table}[t]
\centering
\caption{Pico~2 FP32 On-Device Deployment Summary}
\label{tab:pico2_fp32}
\resizebox{\columnwidth}{!}{%
\begin{tabular}{@{}lccccc@{}}
\toprule
\textbf{Model} & \textbf{Runs} ($\uparrow$) & \textbf{Fails} ($\downarrow$) & \textbf{Latency (ms)} ($\downarrow$) & \textbf{Model (B)} ($\downarrow$) & \textbf{Parity (\%)} ($\uparrow$) \\
\midrule
\modelnameplain{} & \textbf{8} & \textbf{0} & \textbf{83.0} & \textbf{555,221} & \textbf{100.0} \\
TinierHAR & 7 & 1 & 231.5 & 647,156 & 100.0 \\
TinyHAR & 5 & 3 & 486.1 & 626,212 & 89.6 \\
DeepConvLSTM & 1 & 7 & 2034.6 & 1,273,108 & 26.3 \\
\bottomrule
\end{tabular}%
}
\end{table}

\begin{table}[t]
\centering
\caption{ESP32 FP32 On-Device Deployment Summary}
\label{tab:esp32_fp32}
\resizebox{\columnwidth}{!}{%
\begin{tabular}{@{}lccccc@{}}
\toprule
\textbf{Model} & \textbf{Runs} ($\uparrow$) & \textbf{Fails} ($\downarrow$) & \textbf{Latency (ms)} ($\downarrow$) & \textbf{Model (B)} ($\downarrow$) & \textbf{Parity (\%)} ($\uparrow$) \\
\midrule
\modelnameplain{} & \textbf{7} & \textbf{1} & \textbf{67.3} & \textbf{498,528} & \textbf{100.0} \\
TinierHAR & 3 & 5 & 96.4 & 514,784 & 100.0 \\
TinyHAR & 3 & 5 & 352.9 & 517,792 & 88.1 \\
DeepConvLSTM & 0 & 8 & --- & --- & --- \\
\bottomrule
\end{tabular}%
}
\end{table}

\begin{figure}[t]
    \centering
    \includegraphics[width=\columnwidth]{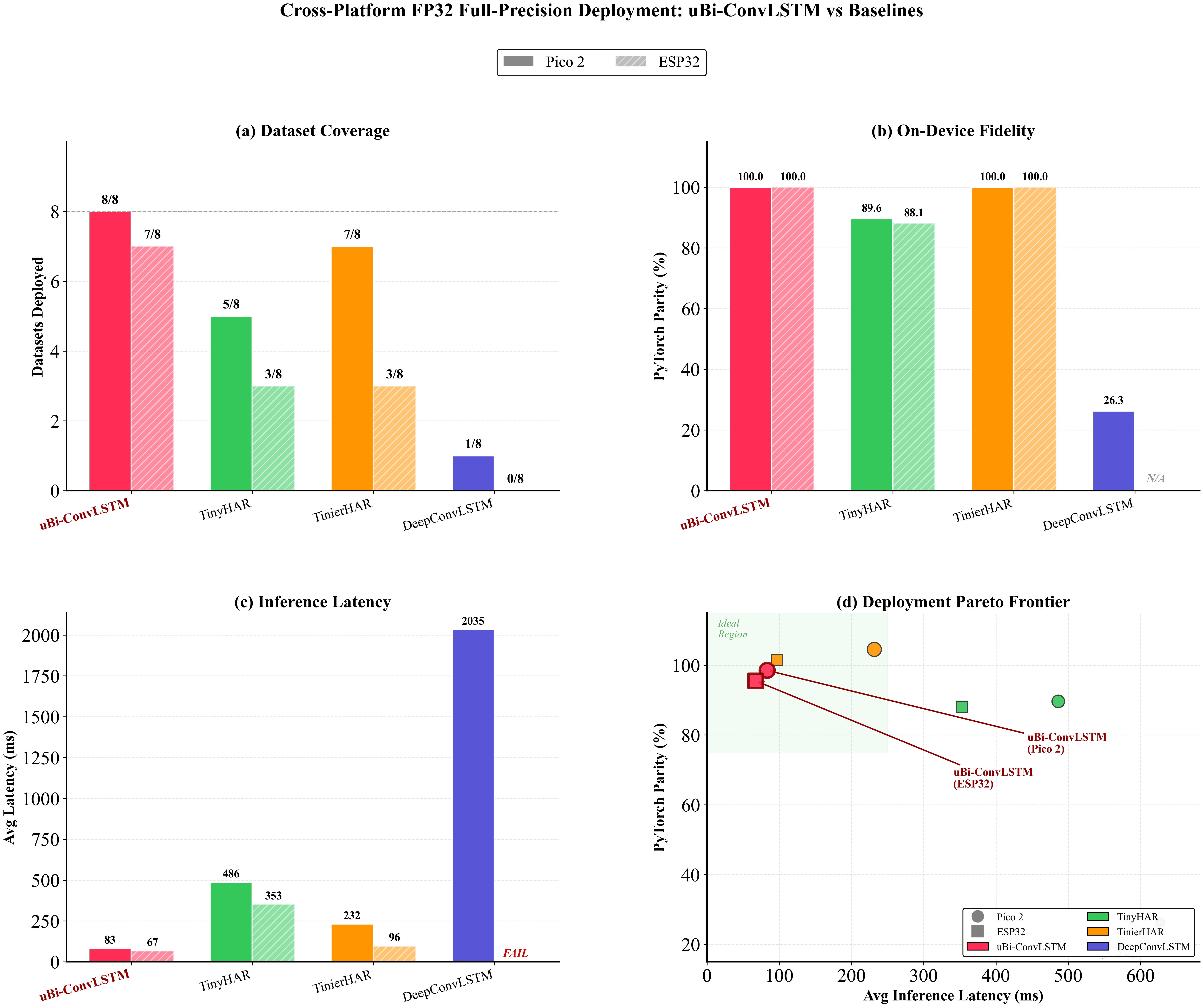}
    \caption{Cross platform FP32 full precision deployment comparison across Pico~2 and ESP32: (a)~dataset coverage, (b)~average PyTorch parity, (c)~average inference latency, and (d)~deployment Pareto frontier.  Hatched bars denote ESP32, and solid bars denote Pico~2.}
    \label{fig:fp32_deployment}
\end{figure}

On Pico~2, \modelname{} again achieves full 8/8 FP32 coverage with 100.0\% PyTorch parity and 83.0~ms average latency, confirming that all INT8 parity degradation observed in Table~\ref{tab:pico2_deployment} is due to quantization rather than to architectural or runtime limitations. TinierHAR deploys on 7/8 datasets (failing only on Opportunity due to buffer resize overflow) with perfect parity on all successful runs, demonstrating that the 54.2\% INT8 parity reported in Section~V-C was entirely a quantization artifact. TinyHAR achieves 5/8 FP32 coverage (failing on PAMAP2, Opportunity, and SKODA) at 89.6\% average parity, where the remaining parity gap on Daphnet (64.9\%) persists even in FP32, and may reflect a training convergence issue rather than a deployment artifact.

On ESP32, \modelname{} achieves 7/8 FP32 coverage (failing only on Opportunity, where the unquantized arena exceeds 185~KB internal SRAM) with 100.0\% parity on all successful datasets, and 67.3~ms average latency, which is 2.5$\times$ faster than the INT8 path on the same platform. This latency reduction occurs because the ESP32's Xtensa core natively supports single-precision floating point operations through hardware, making FP32 inference faster than the software emulated INT8 dequantize-compute path in \texttt{esp-tflite-micro}. TinierHAR and TinyHAR each deploy on only 3/8 FP32 datasets (WISDM, UniMiB, and Daphnet), with TinierHAR achieving perfect parity on its successful configurations. DeepConvLSTM fails on all 8/8 datasets in FP32 as well, confirming that its memory footprint is fundamentally incompatible with both platforms regardless of precision.

The FP32 results provide three key insights. First, the INT8 parity degradation observed in Sections~V-C and V-D is confirmed to be a quantization artifact rather than an architectural deployment limitation, because all models that deploy in FP32 achieve higher fidelity. Second, FP32 coverage is lower than INT8 coverage for every architecture except \modelname{} on Pico~2 (where both achieve 8/8), showing the fundamental coverage fidelity tradeoff in TinyML deployment. Third, \modelname{} is the only architecture achieving near complete FP32 coverage on both platforms (8/8 Pico~2, 7/8 ESP32), confirming that its compact tensor arena footprint provides deployment headroom that larger architectures lack even when quantization overhead is removed. Figure~\ref{fig:fp32_deployment} visualizes the FP32 deployment landscape.

\section{Ablation Studies}

Systematic ablations were conducted, to validate each architectural component. Five variants are evaluated: A0 (Base), A1 (No Pool) removing max pooling layers, A2 (UniDir) using unidirectional LSTM, A3 (Single Conv) removing the second convolutional stage, and A4 (Mean Pool) replacing last timestep aggregation with mean pooling.

\begin{figure*}[t]
    \centering
    \includegraphics[width=0.9\textwidth]{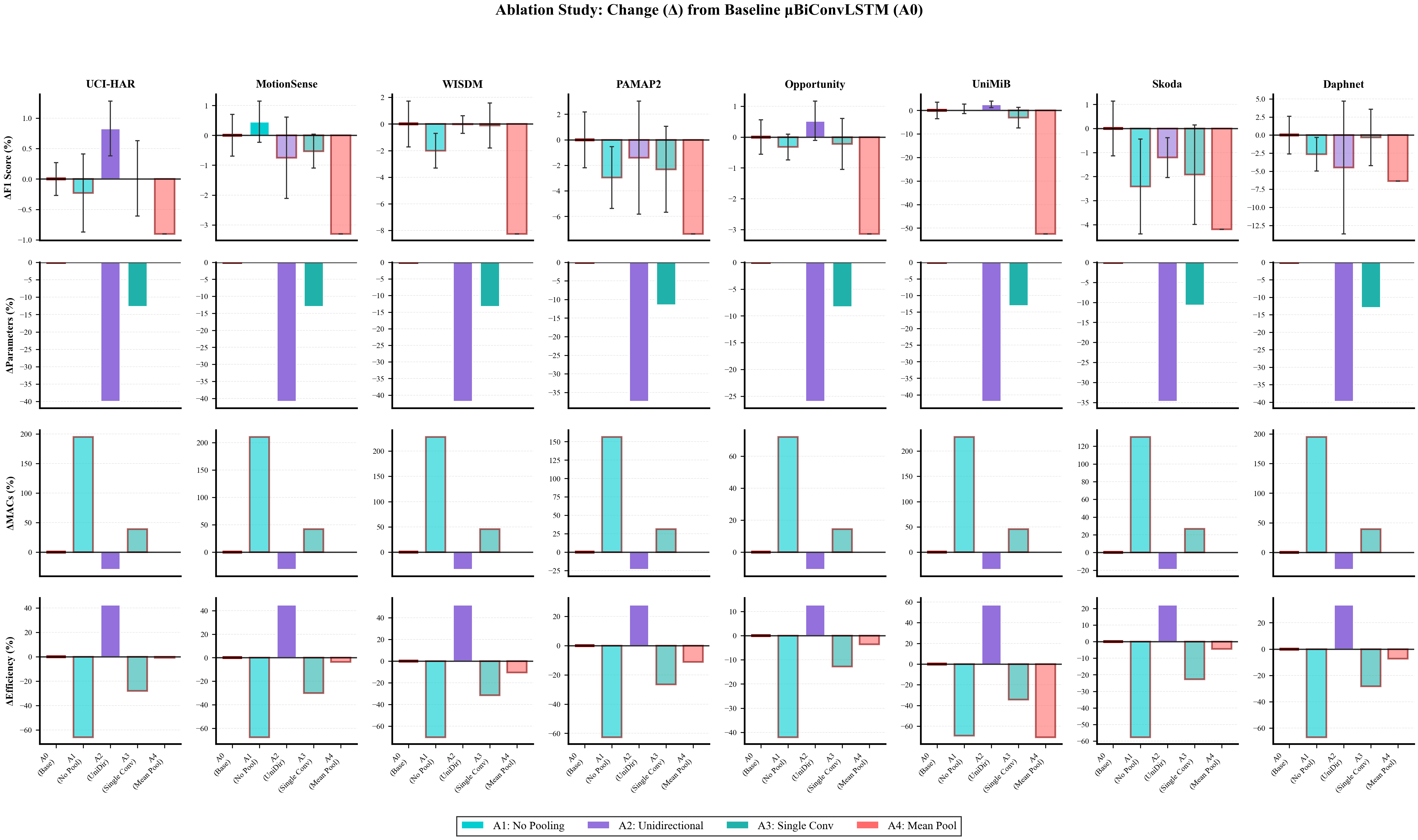}
    \caption{Ablation results across variants (A0--A4) and datasets. A0: Base configuration, A1: No pooling, A2: Unidirectional LSTM, A3: Single conv block, and A4: Mean pooling aggregation.}
    \label{fig:ablation_grid}
\end{figure*}

Figure~\ref{fig:ablation_grid} presents ablation results across all variants and datasets. Several insights emerge.

\textbf{Task Dependent Bidirectionality}: Removing the backward pass (A2) degrades Daphnet by 4.48\%, but \textit{improves} UCI-HAR by 0.83\%. This suggests periodic locomotion activities have sufficient forward temporal structure, while episodic events (freezing) benefit from bidirectional context to capture both onset and recovery patterns. However, this benefit is modest compared to other architectural factors.

\textbf{Important Aggregation Choice}: Mean pooling (A4) causes catastrophic 52.43\% degradation on UniMiB while minimally affecting WISDM ($-$0.21\%). UniMiB's fall detection requires identifying brief transient events where the final bidirectional state captures important endpoint information lost through averaging.

\textbf{Temporal Compression Value}: Removing max pooling (A1) increases MACs by 3.1$\times$ with inconsistent accuracy effects, validating the aggressive temporal compression strategy. The pooling layers provide regularization that benefits generalization on smaller datasets.

\begin{figure}[t]
    \centering
    \includegraphics[width=\columnwidth]{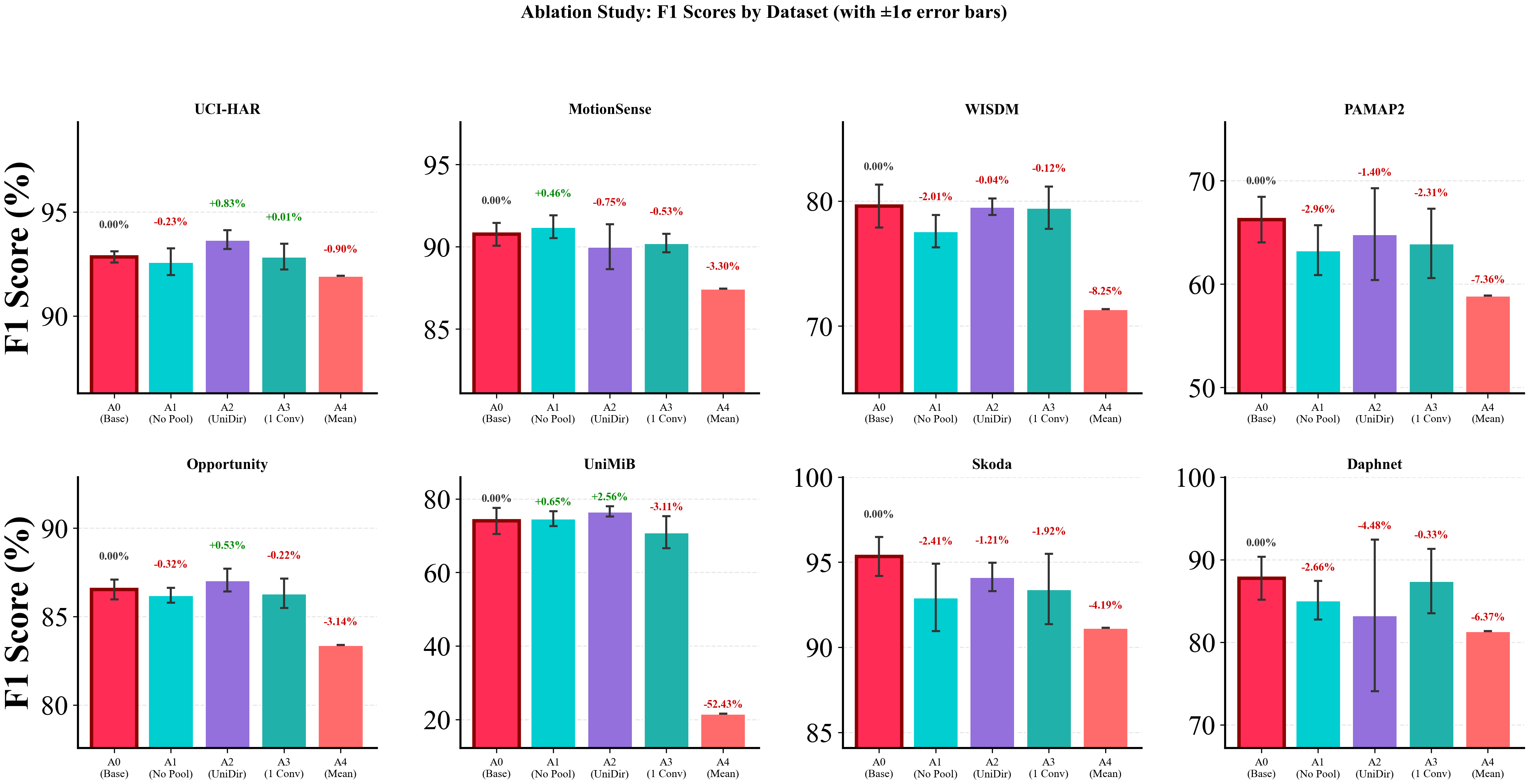}
    \caption{F1-score differences from base configuration (A0) across ablation variants and datasets.}
    \label{fig:ablation_absolute}
\end{figure}

As shown in Figure~\ref{fig:ablation_absolute}, the base configuration achieves best or near best performance on 5 of 8 datasets. Hyperparameter sensitivity analysis (varying convFilters 8-24, lstmHidden 16-32) shows the base configuration (16 filters, 24 hidden) sits on the Pareto frontier. Scaling to 24/32 yields only 0.79\% average F1 improvement across datasets while nearly doubling parameters (from approximately 10.5K to 19.3K), confirming diminishing returns beyond the base configuration.

\subsection{Robustness Analysis}
\label{sec:ablation}

Simulating sensor degradation, we find 0.13\% F1-score degradation when dropping every 5th sample (jitter), but 86.29\% degradation on complete accelerometer dropout, confirming effective sensor specific feature learning rather than trivial signal copying.

\begin{figure}[t]
    \centering
    \includegraphics[width=\columnwidth]{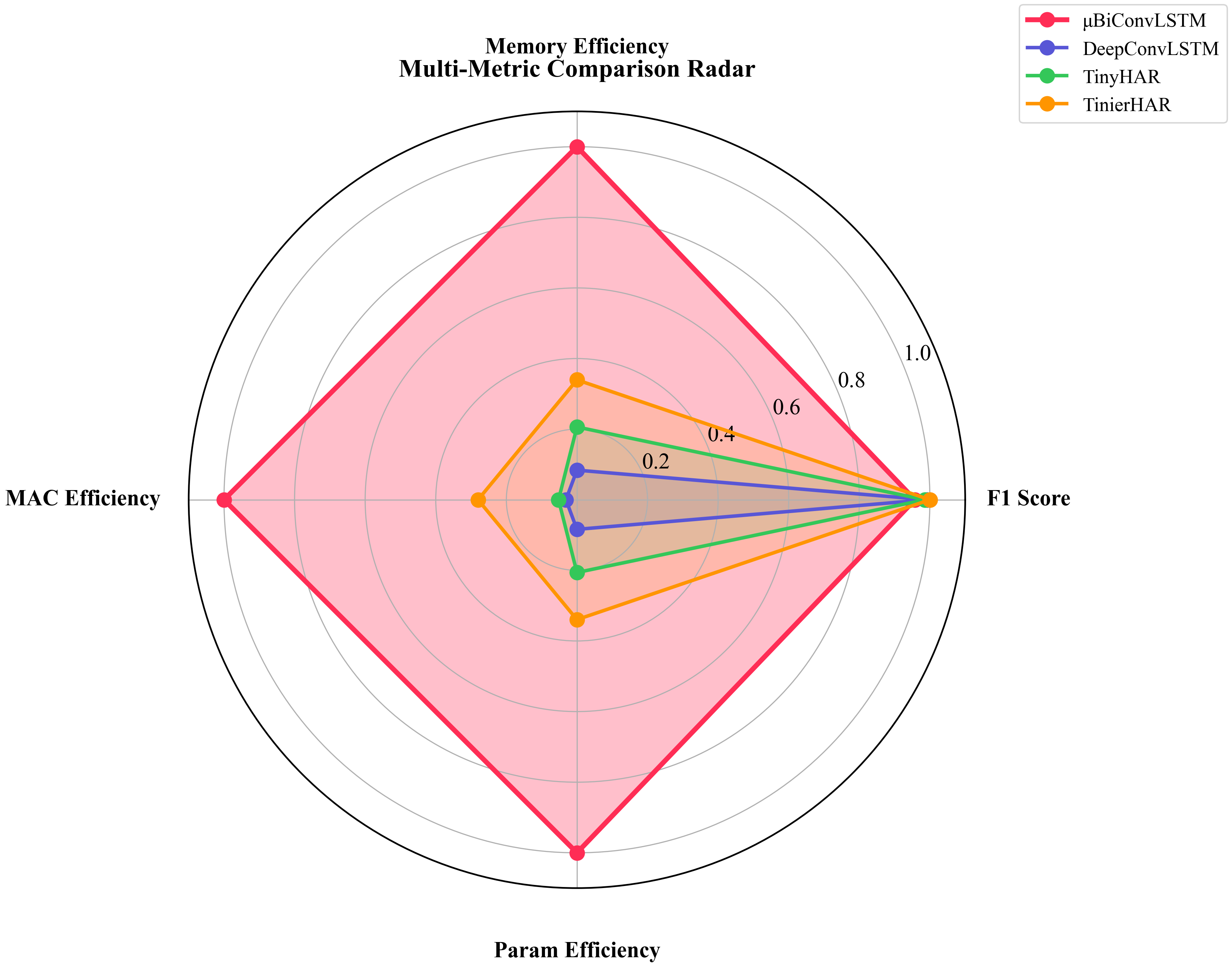}
    \caption{Radar plot comparing four architectures across five normalized metrics: accuracy (mean F1-score), parameter efficiency (F1 per thousand parameters), computational efficiency (F1 per million MACs), memory efficiency (inverse of INT8 model size in KB), and quantization robustness (1 minus degradation percentage).}
    \label{fig:radar}
\end{figure}

Figure~\ref{fig:radar} provides a comparison. All metrics are scaled to [0,1] where larger values indicate better performance. \modelname{} achieves the largest area in efficiency dimensions while accepting modest accuracy tradeoffs, demonstrating suitability for deployment scenarios where memory and computational constraints dominate.

\section{Discussion}

\subsection{Use cases favoring \modelname}

The results reveal that aggressive architectural compression can maintain competitive accuracy when design choices align with task structure:

\textbf{Smartphone IMU Data (3-9 channels)}: On UCI-HAR and MotionSense, \modelname{} achieves within 3\% of larger baselines despite 2.9$\times$ fewer parameters than TinierHAR. The limited channel count aligns well with our standard convolution design principle (P2).

\textbf{Repetitive Industrial Gestures}: SKODA performance (94.46\%) demonstrates that 4$\times$ temporal pooling preserves discriminative features for cyclic assembly tasks where gesture boundaries are well defined.

\textbf{Binary Medical Classification}: Competitive Daphnet performance (88.98\% vs.\ TinierHAR's 89.84\%) suggests bidirectional processing captures gait freeze temporal dynamics effectively, though the margin is modest.

\subsection{Limitations}

\textbf{Class Imbalance Sensitivity}: On PAMAP2, \modelname{} shows a 13.3\% F1-score gap as compared to TinierHAR due to extreme class imbalance where rare activities (rope jumping, cycling) constitute $<$10\% of samples, causing ultra-lightweight models to prioritize majority classes under capacity constraints. Notably, Opportunity achieves competitive performance (87.58\% vs.\ 87.09\%) despite 79 channels, demonstrating that class distribution impacts ultra-lightweight models more than sensor dimensionality. Practitioners working with severely imbalanced datasets should consider larger architectures or specialized sampling strategies.

\textbf{High Variance on Challenging Datasets}: WISDM results show that 12.4\% standard deviation, likely due to the dataset's lower sampling rate (20~Hz), and minimal channel count (3) creating challenging learning dynamics. This suggests sensitivity to data quality in the ultra-lightweight regime.

\textbf{Modest Bidirectionality Benefits}: Ablation studies reveal bidirectionality provides inconsistent benefits across tasks. For periodic activities like walking, unidirectional processing suffices or even improves performance. The parameter cost of bidirectionality (approximately 50\% of LSTM parameters) may not be justified for all applications.

\subsection{Deployment Considerations}

The on-device results across both INT8 and FP32 paths (Sections~V-C through V-E) change the deployment discussion from theoretical projection to empirical evidence. Under INT8 quantization, \modelname{} achieves full 8/8 dataset coverage on both the Pico~2 (72.8~ms average, with 85.7\% parity) and ESP32 (168.5~ms average, with 97.9\% parity). Under FP32 full precision deployment, \modelname{} achieves 8/8 on Pico~2 and 7/8 on ESP32, both with 100.0\% PyTorch parity, confirming that all INT8 fidelity degradation is a quantization artifact. No other architecture achieves full coverage on both platforms under either precision path. DeepConvLSTM is non-deployable on both platforms in both INT8 and FP32 despite strong desktop accuracy.

The FP32 results isolate quantization as the sole source of the parity gaps reported in Sections~V-C and V-D. TinierHAR's INT8 parity on Pico~2 (54.2\%) recovers to 100.0\% in FP32, confirming that the CMSIS-NN INT8 kernel path is the bottleneck rather than any architectural deployment limitation. The corrected ESP32 INT8 baselines add further nuance: replacing the full INT8 export recipe with INT16 activations and INT8 weights recovered considerable fidelity for TinierHAR (90.8\%) and TinyHAR (88.6\%), though the corrected exports sometimes traded fidelity for increased memory, pushing previously deployable configurations past the internal SRAM limit.

The INT8 parity gap between platforms (85.7\% on Pico~2 vs.\ 97.9\% on ESP32) highlights that on-device quantized inference fidelity is backend dependent: the ESP32's Xtensa \texttt{esp-nn} kernels produce more favorable fixed point rounding than the Cortex-M33's CMSIS-NN path. This cross platform variation is an underreported concern in TinyML research, where desktop INT8 evaluation is often treated as a reliable proxy for on-device behavior. The FP32 deployment path eliminates this variation entirely, but at the cost of reduced coverage due to larger tensor arenas.

The natural deployment strategy depends on the application constraint. For maximum coverage with acceptable fidelity, INT8 deployment of \modelname{} provides guaranteed 8/8 coverage on both platforms. For maximum fidelity on platforms with sufficient memory, FP32 deployment of \modelname{} achieves perfect parity on all deployable configurations. For applications where partial coverage is acceptable, TinierHAR in FP32 represents a viable alternative on platforms with sufficient IRAM budget.

\section{Conclusion}

 \modelname{} is an ultra-lightweight HAR architecture achieving 2.9$\times$ parameter reduction as compared to TinierHAR (11.4K vs.\ 34K parameters on average) while maintaining competitive accuracy within the ultra-lightweight regime. Key design choices, namely aggressive 4$\times$ temporal pooling, single-layer bidirectional LSTM, and last timestep aggregation, yield a compact model that deploys successfully on both the Raspberry Pi Pico~2 and ESP32 across all eight benchmarks under INT8 quantization.

On-device validation confirms that \modelname{} is the only architecture in this study achieving full or near full dataset coverage on both target MCUs under both precision paths. Under INT8 quantization, it achieves 8/8 coverage on both platforms with 72.8~ms average latency and 85.7\% PyTorch parity on Pico~2, and 168.5~ms and 97.9\% parity on ESP32. Under FP32 full precision deployment, it achieves 8/8 on Pico~2 and 7/8 on ESP32 with 100.0\% parity on all successful configurations, confirming that all INT8 fidelity degradation is a quantization artifact rather than an architectural limitation. By contrast, DeepConvLSTM (136K parameters) fails to deploy on either platform under both precision paths, while TinyHAR and TinierHAR achieve only partial coverage with considerable parity degradation under INT8 quantization.

Ablation studies reveal that architectural contributions are task dependent: bidirectionality benefits episodic event detection but provides marginal or negative impact on periodic activities. The cross platform INT8 parity discrepancy (85.7\% vs.\ 97.9\%) exposes an underreported concern in TinyML research, namely that desktop INT8 evaluation does not uniformly predict on-device fidelity across MCU backends. The FP32 results eliminate this variation entirely, establishing quantization as the dominant fidelity bottleneck.

These findings suggest that the ultra-lightweight regime (under 15K parameters) is viable for single sensor wearable applications when accompanied by rigorous on-device validation across both quantized and full-precision paths. Future work will investigate Neural Architecture Search for sensor-modality-specific configurations, knowledge distillation to recover INT8 parity gaps, and systematic characterization of backend dependent quantization fidelity across additional MCU targets. Code with PyTorch models, and quantized models: https://github.com/WhiteMetagross/MicroBiConvLSTM

\bibliographystyle{IEEEtran}

\end{document}